\documentclass{IOS-Book-Article}    
\usepackage{subfig}

\usepackage{subfig}
\usepackage{graphicx}
\usepackage{indentfirst}
\usepackage{latexsym}
\usepackage{multirow}
\usepackage{tabls}
\usepackage{wrapfig}
\usepackage{longtable}
\usepackage{supertabular}
\usepackage[rflt]{floatflt}  
\usepackage{colortbl}  
\usepackage{epstopdf}
\usepackage{fixltx2e}
\usepackage{verbatim}
\usepackage{multirow}
\usepackage{amsmath}
\usepackage{algorithm}
\usepackage[noend]{algpseudocode}
\usepackage{amsmath}
\usepackage{wrapfig}
\usepackage{grffile}
\usepackage[table]{xcolor}
\usepackage{booktabs}
\usepackage{float}
\usepackage{times,amsmath,amsfonts}
\usepackage{soul}
\usepackage{makecell}
\usepackage{fancyvrb}

\begin{document}


\begin{frontmatter}

\title{Ontology-Grounded Topic Modeling\\for Climate Science Research\thanks{This is a preprint of work that will be presented at the Semantic Web for Social Good Workshop of the International Semantic Web Conference, October 2018 and published as part of the book "Emerging Topics in Semantic Technologies. ISWC 2018 Satellite Events", E. Demidova, A.J. Zaveri, E. Simperl (Eds.), ISBN: 978-3-89838-736-1, 2018, AKA Verlag Berlin. Copyright held by the authors.}}

\pdfinfo{
/Title (Ontology-Grounded Topic Modeling for Climate Science Research)
/Author (Jennifer Sleeman, Tim Finin, Milton Halem)}

\setcounter{secnumdepth}{0}  

\author[A]{\fnms{Jennifer} \snm{Sleeman}}
\author[A]{\fnms{Tim} \snm{Finin}}
\author[A]{\fnms{Milton} \snm{Halem}}
\address[A]{University of Maryland, Baltimore County, Baltimore, MD 21250 USA\\ 
\{jsleem1, finin, halem\}@umbc.edu} 

\runningauthor{J. Sleeman et al.}

\maketitle           

\begin{abstract}
In scientific disciplines where research findings have a strong impact on society, reducing the amount of time it takes to understand, synthesize and exploit the research is invaluable.  Topic modeling is an effective technique for summarizing a collection of documents to find the main themes among them and to classify other documents that have a similar mixture of co-occurring words. We show how grounding a topic model with an ontology, extracted from a glossary of important domain phrases, improves the topics generated and makes them easier to understand. We apply and evaluate this method to the climate science domain.  The result improves the topics generated and supports faster research understanding, discovery of social networks among researchers, and automatic ontology generation.
\end{abstract}

\end{frontmatter}

\section{Introduction}

The authoritative source for conveying the latest climate research findings, recommendations and mitigations steps is the Intergovernmental Panel on Climate Change (IPCC) \cite{IPCC}.  The reports produced by the IPCC are published every five years and are composed of four separate volumes: Physical Science Basis; Impacts, Adaptations and Vulnerability; Mitigation of Climate Change; and Synthesis Reports. Each IPCC volume has eight to twenty-five chapters and each chapter cites between 800 and 1200 external research documents.  The IPCC reports provide not only a comprehensive assessment of the climate science, but the analysis of the 30-year series of reports shows how the scientific field has and continues to evolve.

For a new climate scientist, absorbing this information in order to perform research or make policy contributions can be daunting.  However, if machine understanding of these reports could be used to summarize, synthesize and model the knowledge, the researcher's task is improved.  We propose this could improve the overall scientific research contributions that could be made by making the process more efficient. 

In our previous work \cite{sleemanPhd,sleeman2016dynamic,sleeman2017modeling,8258063} we described a process by which we converted 25 years of IPCC reports and their cited articles into raw text.  We treated these two document collections, the report chapters and the scientific research papers they cite, as two different \textit{domains}.  We used a topic modeling cross-domain approach to show how these two domains interacted and how the cited research in one report influenced the subsequent reports.  This allows us to more accurately predict how the field of climate science is evolving.

We quickly discovered that the standard topic modeling approaches did not work as well as we hoped on the text in the report domain, and were even less effective on the cited research documents from the scientific domain.  One reason is that scientific literature is written more formally and typically contains more phrases that provide the context through which one understands the literature \cite{fang2005scientific}. Many phrases in a given scientific domain do not follow the usual pattern of compositional semantics in which their meaning can be obtained by combining the meanings of their words.  Rather, they have a specific meaning in the domain that must be learned.  In climate science, for example, \textit{black carbon} refers not to carbon whose color is black, but to the sooty material emitted from gas and diesel engines, coal-fired power plants and other sources that burn fossil fuel.

\section{Background}

Topic modeling has a long history of relevance to natural language processing, often used to model large collections of text documents applied to problems such as document summarization \cite{blei2003latent}, classification \cite{mcauliffe2008supervised,ramage2009labeled}, recommendation \cite{wang2011collaborative} and search \cite{rosen2004author}. The method, Latent Dirichlet Allocation (LDA), \cite{blei2003latent} made a significant footprint in natural language research.  In Latent Dirichlet Allocation (LDA) \cite{blei2003latent,blei2012}, every document is assumed to be a mixture of topics represented as a probability distribution, and each topic is a probability distribution over the terms in the vocabulary that is formed from the full collection of documents.  Topics are drawn from a Dirichlet distribution. 

Topic modeling is often used to find word vectors that best represent the themes in a collection of documents.  Each word vector contains a set of words or word phrases, where each word/word phrase has an associated probability that represents that word or word phrase's contribution in representing a particular theme.  The approach most frequently used is to extract words from documents, removing commonly occurring words and symbols, and to generate a `collection vocabulary' from the set of words.  Sometimes the vector is a set of singleton words, however, word n-grams, where the `n' represents the number of words that make up the phrase, are also used.  

It has been shown that when word phrases are used in topic models, the topics tend to be more relevant to the collection of documents \cite{wallach2006topic,lindsey2012phrase,wang2013phrase,el2014scalable}.  We have found this to be particularly important when the collection of documents pertains to a scientific discipline \cite{sleeman2017modeling}. However, knowing what sort of phrases one should extract and use to train the topic model is a problem.  The standard bag of words approach is often used, where each word from the document is treated as a singleton. 

\section{Topic Modeling for Scientific Domains}
Particularly for scientific domains, phrases often convey special meaning that is lost by treating them as single words.  The challenge with using word n-grams is knowing which sequences carry such meaning, making automatic phrase or word n-gram extraction problematic.  The fact that key phrase extraction is currently an active areas of research among \cite{mahata2018key2vec,yu2018wikirank}, it is clear this is still a challenging problem.

To understand this challenge further, we used tools from NLTK \cite{nltk} to find common, meaningful phrases that align with concepts in a space science domain (e.g., \textit{'active galactic nucleus'}) but other, less relevant ones were also found, such as \textit{'aboard the Hubble'} and \textit{'central region of'}.  Stop word removal can filter some, but not all of these words and many n-grams would require human judgment to filter.  In this example, there were also missed phrases such as \textit{'Chandra X-Ray Observatory'}, where instead \textit{'Ray Observatory'}, \textit{'the observatory'}, and \textit{'- Ray Observatory'} were found.

Instead we ground the topic modeling process on a domain-specific ontology seeded with predefined key word phrase concepts obtained from domain-specific sources such as domain experts, and by data mining semi-structured sources. In particular, we found the IPCC glossaries and domain experts to be good sources for defining climate-related word phrase concepts.  This grounding process contextualizes the topic model such that the topics are more relevant to the domain that is being modeled.  For example, given a climate change domain ontology, if a document being used to train the topic model included text unrelated to climate change, those words would potentially have a lower weight than the words which represented the 'known' or 'seed' concepts found in the ontology.  

Table \ref{table:glossary} shows examples from the data-mined and a bi/trigram extractor approach. Using a sample of climate change data, all bigrams and trigrams were annotated.  The documents were processed using the data mining approach, and also using an n-gram extraction approach. The extractor approach was only able to recover 6\% of the word phrases that could be 100\% represented using our data mining of glossaries approach. Similar results were found for space science data.  Though more recent research \cite{mahata2018key2vec,yu2018wikirank} may significantly improve upon a typical bi/trigram extractor, for scientific data, seeding the ontology with known concepts that are readily available through published glossaries provides the more accurate set of concepts.

\begin{table}[p]
\centering
\caption{Examples of data-mined phrases and bigram/trigram extracted phrases}
{

\label{table:glossary}
\label{table:extractor}
\normalsize{}
\begin{tabular}{|p{4in}|} \hline
\rowcolor[gray]{0.9}
\emph{Example Word Phrases Mined From Glossaries} \\ \hline
Ultraviolet Radiation (UV)\\
Forcing Mechanism\\
Fossil Fuel\\
Water Vapor\\
General Circulation Model (GCM)\\
Fluorinated Gases\\
United Nations Framework Convention on Climate Change (UNFCCC)\\
Feedback Mechanisms\\
100-Year Flood Levels\\
\hline
\end{tabular}

\vspace{0.25in}
\begin{tabular}{|p{4in}|} \hline
\rowcolor[gray]{0.9}
\emph{Example Automatic N-Gram Extractions Using NLTK}\\ \hline
 World Meteorological Organization\\
 General Circulation Models\\
 the atmosphere that\\
 in the Earth\\
 climate system \\
 Assessment Report Working\\
\hline
\end{tabular}
}
\end{table}

\begin{figure}[p]
\centering
\fbox{\includegraphics[width=.8\textwidth]{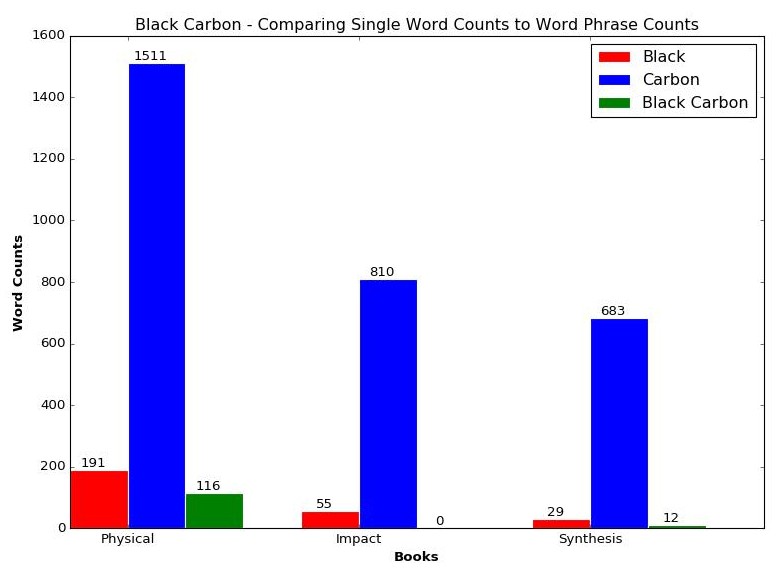}}
\caption{Comparing Word Counts and Phrase Counts for Black Carbon Among IPCC Books Assessment Report 3.}
\label{fig:black_carbon}
\end{figure}

Glossaries typically provide key concepts that are relevant to a particular domain and consist of words and phrases whose length can be anywhere from two to ten words.  For example, in the climate change community, the phrase \emph{"soil moisture'} implies something much more meaningful than 
\emph{`soil'} and \emph{`moisture'} alone.  Furthermore,  the singleton words might be much more frequently found than the phrase. This is often an important artifact in the topic model.  For example, \emph{`black carbon'} is a significant concept in climate  change because of its impact on the research at a certain period of time.  The word \emph{`black'} may not frequently occur with other words but among climate change literature the word \emph{`carbon'} occurs quite frequently with other words.  An example of this is shown in Figure \ref{fig:black_carbon}, where the phrase \emph{`black carbon'} has a significantly lower occurrence across the Physical Science, Impact and Synthesis books for Assessment Report 3 than the single word \emph{`carbon'}.  The single word \emph{`black'} not only appears within the phrase \emph{`black carbon'} but also within the phrase \emph{`black spruce'}, \emph{`black-footed ferrets'}, and within other phrases.

\section{Approach}

\begin{figure}[H]
\centering
\fbox{\includegraphics[width=.45\textwidth]{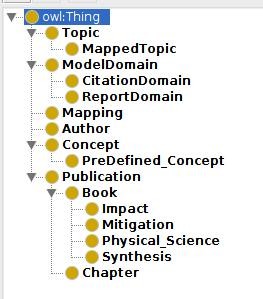}}
\caption{A Partial IPCC Ontology Used for Guiding Topic Modeling.}
\label{fig:onto_ex}
\end{figure}

\begin{figure}[H]
  \centering
  {\fbox{\includegraphics[width=\textwidth]{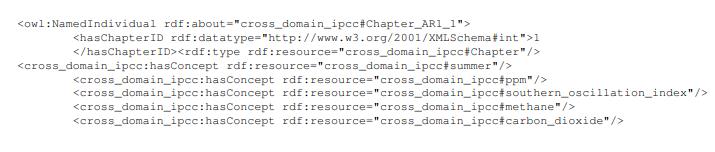}}}
  \caption{An Example of Chapter Concepts Ontologically Represented.}
  \label{fig:concepts_ex1}
\end{figure}

\begin{figure}
 \centering
   \fbox{\includegraphics[width=\textwidth]{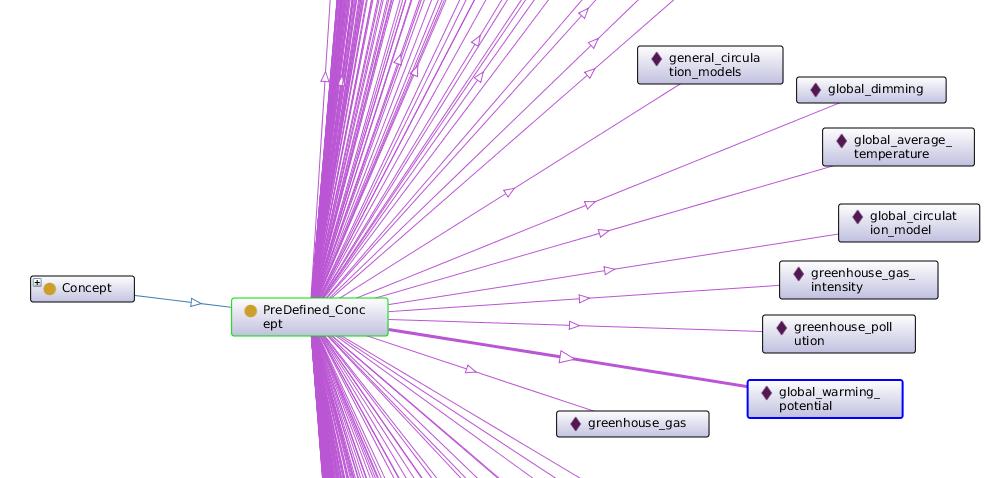}}
      \caption{An Example of Ontology Concepts from the IPCC Ontology.}
   \label{fig:concepts_ex2}
\end{figure}

Our approach entails modeling the structure of the reports and their citations in the ontology. There were five IPCC assessment reports, AR1-AR5, each of which follows a similar structure consisting of four distinct books: Physical Science Basis, Impacts, Adaptations and Vulnerability, Mitigation of Climate Change, and Synthesis Reports. Each book has between 11 and 25 chapters and a chapter typically cites between 800 and 1200 external documents.  The ontology consists of a similar structure as shown in Figure \ref{fig:onto_ex}. We then obtain a list of concepts of importance from domain experts and domain glossaries.  Figures \ref{fig:concepts_ex1} and \ref{fig:concepts_ex2} shows example pre-defined seed concepts represented in our IPCC ontology

In addition, acronyms are mapped to the actual phrase and treated as the same concept. For example, \emph{`ENSO'} is treated as the the same concept as \emph{`El Nino Southern Oscillation'}.  The ontology is then read into memory for the preprocessing step of the topic modeling phase. As we perform preprocessing of text, we use the ontology concepts for weighting concepts we find in the text.  We do this for both the report data and the citation research papers.  This retrieval process is described in more detail in \cite{sleeman2017modeling}.  After we perform the topic modeling phase, we update the domain ontology with the concepts associated with each chapter of the book, the topics generated with word probabilities, and cross-domain mappings between the reports domain and the research paper domain.  Figures \ref{fig:concepts_ex1} and \ref{fig:concepts_ex2} shows a small subset of the concepts extracted from the First Assessment Period, Chapter 1.

\subsection{Ontologically Represented Topics}
\begin{figure}
  \centering
  \fbox{\includegraphics[width=\columnwidth]{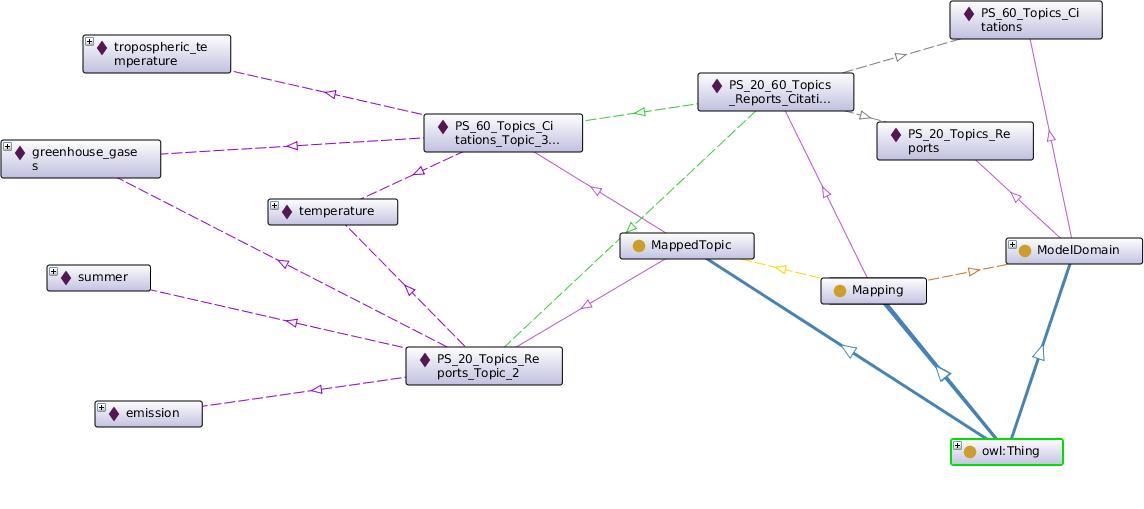}}
    \caption{Example of an Ontologically Represented Topic Mapping Between Two Topic Domains.}
  \label{fig:mt1}
\end{figure}

Since ontological word phrases are used to ground the topic modeling process, topics are also represented by an ontological structure.  Topics are implicitly linked into the ontological structure, along with documents.  The IPCC reports have citation information for each chapter in the report. In our work we built a topic model for the citations and another topic model for the reports.  We used the ontology as a means for conveying how the two topic models were related to each other using the topics generated from the two different models as a bridge between the two models.  For example, Figure \ref{fig:mt1} shows a set of common concepts from two mapped topics captured ontologically.

\subsection{Ontologically Represented Social Networks}
Since we captured the citation information ontologically, we can use this information to discover interesting social communities.  For example, a class in the ontology called `Publication' can have many `Authors'.  When a number of authors are referenced together across multiple publications, this can give insight into a social relationship between these authors.  Specifically, if the same authors are cited in three different chapters that relate to \emph{`Black Carbon'}, these authors form a relationship with a common node  concept \emph{`Black Carbon'}.  The ontology can also be used to observe social networks given the relationships between authors that cite authors that are also cited in the same chapter.  

\begin{figure}
\centering
   \fbox{\includegraphics[width=\textwidth]{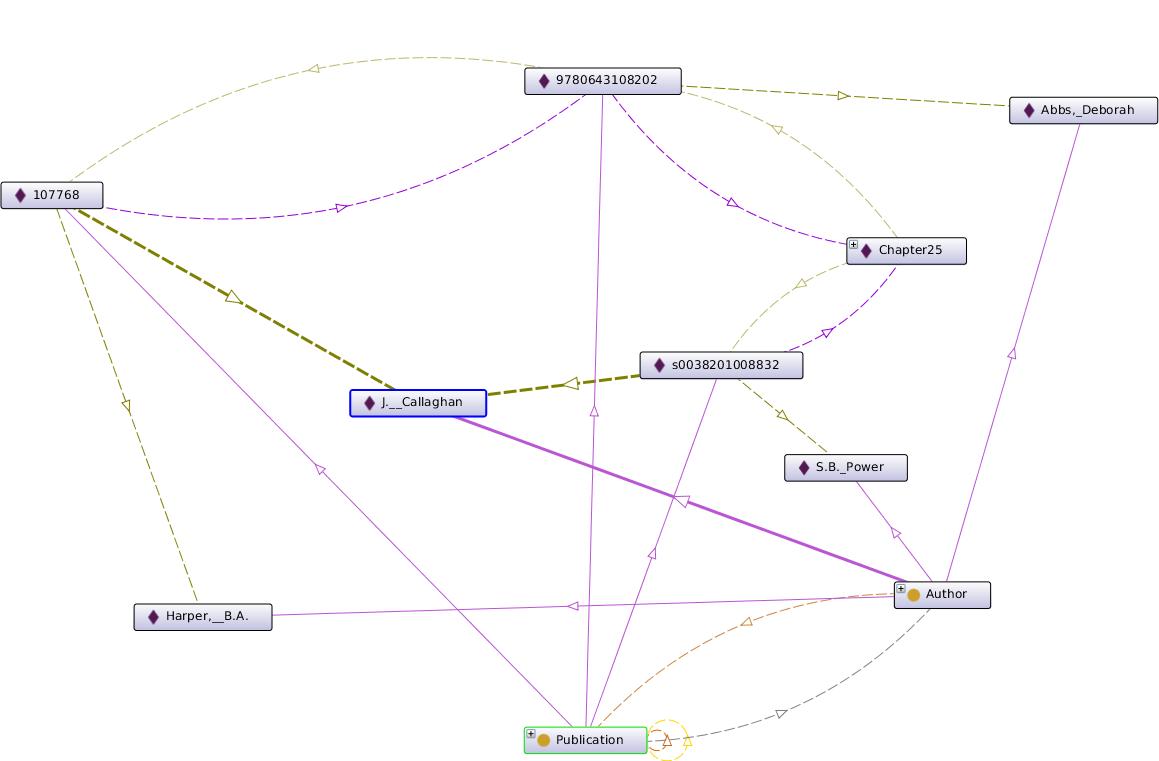}}
   \caption{Example of a Social Network from Citation Connections.}
   \label{fig:mt2}
\end{figure}

Figure \ref{fig:mt2} shows an example in which \emph{`Callaghan'} was an author in a paper cited in one chapter and author \emph{`Abbs'} was also cited in the same chapter. 
     
Of the paper for which \emph{`Abbs'} is an author, a relationship was found between \emph{`Abbs'} and \emph{`Callaghan'} due to the fact that they were both cited in the same chapter and one cited the other in their cited paper.  The ontological representation for citations can also shed light, in general, on which authors are cited across books and chapters.

\section{Experimentation and Evaluation}

We used our ontology-guided topic modeling approach to model both the IPCC reports and the citations.  We first converted the reports and citations to raw text, we extract meta-information regarding the citations and we used the ontology to link this information together.  When a predefined concept from our ontology was found, we weighted the phrase higher than non-ontology concepts by 5\%, 10\%, 25\%, and 50\%. To understand how ontological concept-based topic modeling differs from standard bag of words topic modeling, topic models were built using both approaches.  The ontological grounded topic model was compared to a bag of words model that used the same data set, the same stop word removal, but without the ontology concepts grounding the modeling.  Therefore it did not contain word phrases but could potentially contain word phrases as individual words.  Perplexity was used to evaluate the two models.  

In this experiment, the IPCC `Physical Science' book for assessment periods one through five were used. There were 61 documents in total used in this topic model, with 11 documents in AR1, 11 documents in AR2, 14 documents in AR3, 11 documents in AR4, and 14 documents in AR5.  The assessment reports were used for this experiment and each chapter was treated as a document.  For perplexity evaluations, a held-out set was used for each assessment period beginning with AR2. One way to understand the differences between these two models is by simply observing the topics.  

\begin{table}
\centering
\caption{Example Topic Output Comparing the Bag-of-Words and Ontology-Grounded Models Using the Top 10 Terms.}
{
\label{table:ex_bow_ogtm1}
\begin{tabular}{|p{2.25in}|p{2.25in}|}\hline
\rowcolor[gray]{0.9}
\textbf{Bag of Words Topics} & \textbf{Ontology-Based Word Phrases Topics} \\ \hline
greenhouse, effect, forcing, warming, temperature, global, detection, signal, variability, pattern
 &
radiative, radiative forcing, effect, greenhouse gases, carbon dioxide, emission, concentration, ozone, aerosol, solar
 \\\hline
 ocean surface, global, simulation, cloud, system, temperature, atmosphere, atmospheric, heat
&
 el nino southern oscillation, ocean, global, temperature, land, sea level rise, estimate, sea surface temperature, precipitation, cloud
\\
\hline
\end{tabular}
}
\end{table}

For example, in Table \ref{table:ex_bow_ogtm1} the two topics highlighted appear to be the `radiative forcing' topic for each model.  The ontology guided model provides a richer set of words since the topic is composed of phrases.  The same is also true for the `El Nino' topic for each model.  Though both models provide visually relevant topics, the ontology-guided model provides concepts that are more specific to the scientific domain.  

\begin{figure}[htb]
\centering
\fbox{\includegraphics[width=\textwidth]{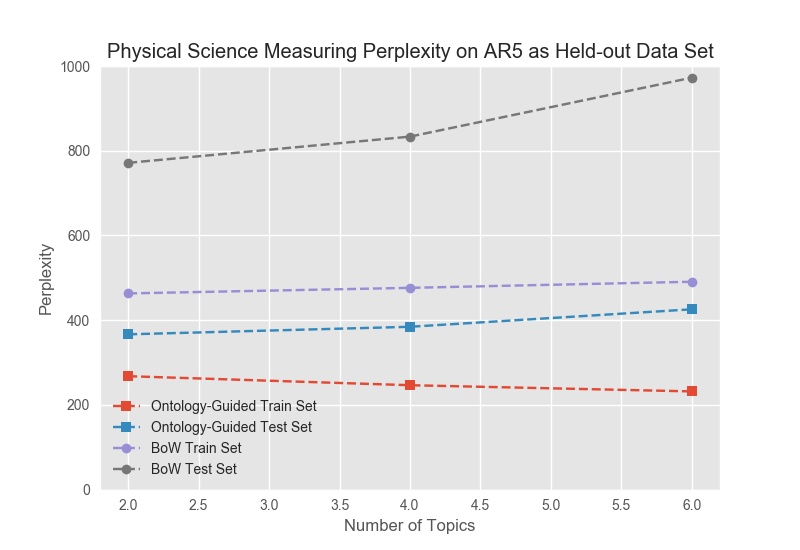}}
\caption{Perplexity of AR5 Held-out Set, Comparing Ontology-Grounded and Bag-of-Words Models.}
\label{fig:ggm_3d}
\end{figure}

A common metric used to evaluate topic models is perplexity.  Perplexity measures how well a probability distribution predicts a held-out sample.  A lower perplexity indicates the model is better at prediction. Perplexity was measured across assessment periods where $t$ is the held-out set of documents, using a  model trained on $t-1$ assessment documents, given $t$ ranges from 2 to 5. Each experiment compared the ontology-grounded and non-ontology-grounded methods. In Figure \ref{fig:ggm_3d}, AR4 was used to build the topic model and AR5 was used as the held-out test set.  Similar perplexity was measured for the other assessments.

Given the size of this data set, the number of topics that best represents this data set is between two and six topics.  This was confirmed by visualizing the topics.  When the topic size grows too large, topics tend to overlap much more.  The difference in perplexity for the same held-out data set is shown, with perplexity measures (lower is better) indicating the ontologically grounded topic modeling method may improve perplexity which in turn means it may offer better predictability for scientific data.  As a mean of reference, the training set is also used as a held-out set so as to show the difference in scale between the ontological method and the non-ontological method.  This provides a general idea as to the performance of these two approaches.

We performed experiments to compare the ontologically-grounded word phrase approach with a standard approach, both of which use the same method for stop word removal. With each experiment the top N words are used for a given set of topics.  Given a second topic model example comparing the ontology-grounded model and the non-ontology grounded model, as shown in Table \ref{table:ex_bow_ogtm2}, the topics in the ontology-grounded model are more closely related to terminology found in scientific research papers when compared with the non-ontology grounded model.  For example, the first four words `temperature', `anthropogenic', `carbon dioxide', and `radiative forcing' are more descriptive than `change', `ocean', `level', and `global'.

\begin{table}[H]
\centering
\caption{A Second Example Showing the Topic Output of the Ontology-Grounded Model Using Top 10 Terms is More Descriptive Than the Bag-of-Words Topic Output.}
{
\centering
\label{table:ex_bow_ogtm2}
\begin{tabular}{|p{2.25in}|p{2.25in}|}\hline
\rowcolor[gray]{0.9}
\emph{Bag of Words Example Topics} & \emph{Ontology-Based Word Phrases Example Topics} \\ \hline
change, ocean, level, global, model, mean, climate, figure, rise, surface &
temperature, anthropogenic, carbon dioxide, radiative forcing, sea level rise, greenhouse gases, snow surface temperature, wind, global warming potential \\\hline
carbon, climate, change, emission, atmospheric, ocean, model, university, global, land &
carbon dioxide, carbon cycle, atmospheric co2, anthropogenic, temperature, land use, methane, fossil fuel, ppm, surface temperature\\
\hline
\end{tabular}
}
\end{table}

Using Google to search on the combination of words, two different sets of documents (examining the top three documents) are returned.  With the word phrases contained in [`temperature', `anthropogenic', `carbon dioxide', `radiative forcing'] the top documents included a Wikipedia page related to `Radiative Forcing' and two IPCC report chapters, plus a number of Google Scholar research paper suggestions.  With the four words contained in [`change', `ocean', `level', `global'], the top three results included pages related to `sea level rise', the first hosted by NASA, a second page on the same concept hosted by NOAA, and the third hosted by EPA. There were no Google Scholar suggestions.  This further supports the assertion that by grounding the topic model with concepts from the ontology, the topics created are more context-specific and hence more fine-grained than the standard approach.  For scientific data, this an important point, as this level of detail provides the context needed to really understand scientific documentation.

\section{Related Work}

Typically topic models use a bag-of-words approach and more recently 1-hot encoded bags of words, leaving it to the implementer to decide what that bag of words contains.  Since the early 2000s, research has focused on ways of improving topic modeling by adding context \cite{mcauliffe2008supervised}, labeled topics \cite{ramage2009labeled}, and phrases \cite{wallach2006topic,lindsey2012phrase,wang2013phrase,el2014scalable}.  Early research explored ways of using word n-grams \cite{wu1999combining} to improve different tasks such as text processing, classification,  
named entity recognition and knowledge base population.  The idea of discovering word phrases in topic modeling was proposed by Wang et al. \cite{wang2007topical} in 2007 and was based on using n-grams in topic models which was proposed by Wallach \cite{wallach2006topic}.  Later work followed that proposed extensions to identifying topical phrases \cite{lindsey2012phrase,wang2013phrase,el2014scalable}. 

Early research explored ways of using word n-grams \cite{wu1999combining}
to improve different tasks such as text processing, classification,  
named entity recognition and knowledge base population.  It is reasonable to believe word n-grams would produce better topics and research has shown this to be true. This idea of discovering word phrases in topic modeling was proposed by Wang et al. \cite{wang2007topical} in 2007 and was based on using n-grams in topic models which was proposed by Wallach \cite{wallach2006topic}.  Later work followed that proposed extensions to identifying topical phrases \cite{lindsey2012phrase,wang2013phrase,el2014scalable}. Work by Jameel et al. in 2013 \cite{jameel2013n} combined n-gram models with temporal documents and was foundational in using ontological concepts to ground the topic modeling process.

Recent developments in topic modeling have started exploring its applicability to scientific concepts. Hall et al. \cite{hall2008studying}, address how scientific ideas have changed over time by modeling temporal changes employing DTM,with probability distributions for the ACL Anthology, a public repository of all papers in the Computational Linguistics journals, conferences and workshops. Their work proposes extensions to their model by integrating topic modeling with the citations as done in this paper.  Tang et al. \cite{tang2015can} investigate the use of topic modeling to identify extreme events based on numerical atmospheric model simulations. They associate text terms with statistical ranges of numerical variables. 

\section{Conclusions, Insights and Future Work}

More recent work  \cite{mahata2018key2vec,yu2018wikirank} related to key-phrase identification could be used in conjunction with domain glossaries to automatically populate the ontology. Since many scientific domains define glossaries as part of the document collection, using a heuristic to parse the glossaries is both feasible and effective for constructing ontology concepts. An ontology-grounded word phrase approach for topic modeling results in topics that contain word phrases, which better represents the scientific information.  Perplexity measures support the ontology-grounded method for this specific IPCC scientific data set use case.  The added benefit of guiding this process with an ontology is that the topics and documents are linked to an ontological representation which could be used to support knowledge base population and question answering systems for climate scientists.  This approach turns the simple bag-of-words topic modeling approach into a powerful knowledge understanding tool.

We plan to apply this technique for other domains to get more experience and further test and evaluate the idea.  We are collecting glossaries and concept lists for the cybersecurity domain and plan to develop topic models using them.  We also hope to explore their use to enhance word embeddings.

\section*{Acknowledgement}

This work was partially supported by a grant of computational resource services from the Microsoft \textit{AI for Earth} program and a gift from the IBM \textit{AI Horizons Network}.

\bibliographystyle{splncs04}
\bibliography{2018}

\end{document}